\documentclass[final,twoside]{IEEEtran}
\usepackage{amsmath,amsfonts}
\usepackage{algorithmic}
\usepackage{algorithm}
\usepackage{array}
\usepackage[caption=false,font=normalsize,labelfont=sf,textfont=sf]{subfig}
\usepackage{textcomp}
\usepackage{stfloats}
\usepackage{url}
\usepackage{verbatim}
\usepackage{graphicx}
\usepackage{cite}
\usepackage{hyperref}
\usepackage{bm}
\usepackage{booktabs}
\usepackage{multirow}
\usepackage{multicol}
\usepackage{color}
\usepackage{soul}

\begin{document}

\title{Sparse Point Clouds Assisted Learned Image Compression}

\author{Yiheng Jiang, 
Haotian Zhang,
Li Li, \IEEEmembership{Member, IEEE,}
Dong Liu, \IEEEmembership{Senior Member, IEEE} \\
and Zhu Li, \IEEEmembership{Senior Member, IEEE}
\thanks{
Y. Jiang, H. Zhang, L. Li, and D. Liu are with the MOE Key Laboratory of Brain-Inspired Intelligent Perception and Cognition, University of Science and Technology of China, Hefei 230037, China (e-mail: bitaswood@mail.ustc.edu.cn; zhanghaotian@mail.ustc.edu.cn; lil1@ustc.edu.cn; dongeliu@ustc.edu.cn).
Z. Li is with the Dept of Computer Science and Electrical Engineering, University of Missouri, Kansas City (e-mail: lizhu@umkc.edu).}}


\markboth{Submitted to IEEE Transactions on Circuits and Systems for Video Technology}%
{Jiang \MakeLowercase{\textit{et al.}}: Sparse Point Clouds Assisted Learned Image Compression}


\maketitle

\begin{abstract}
In the field of autonomous driving, a variety of sensor data types exist, each representing different modalities of the same scene. Therefore, it is feasible to utilize data from other sensors to facilitate image compression. However, few techniques have explored the potential benefits of utilizing inter-modality correlations to enhance the image compression performance. In this paper, motivated by the recent success of learned image compression, we propose a new framework that uses sparse point clouds to assist in learned image compression in the autonomous driving scenario. We first project the 3D sparse point cloud onto a 2D plane, resulting in a sparse depth map. Utilizing this depth map, we proceed to predict camera images. Subsequently, we use these predicted images to extract multi-scale structural features. These features are then incorporated into learned image compression pipeline as additional information to improve the compression performance. Our proposed framework is compatible with various mainstream learned image compression models, and we validate our approach using different existing image compression methods. The experimental results show that incorporating point cloud assistance into the compression pipeline consistently enhances the performance.
\end{abstract}


\section{Introduction}

\label{sec:introduction}

\IEEEPARstart{I}{n} autonomous driving scenarios, vehicles typically carry different sensors that simultaneously collect large amounts of data. Among all of this sensor data, camera images are particularly crucial, playing a significant role in tasks related to autonomous driving such as object detection and semantic segmentation. However, the substantial amount of data generated by cameras occupies a significant amount of storage space and transmission bandwidth. Thus, effectively compressing camera images to conserve resources is a worthwhile research topic. 

Unlike typical image compression scenarios, autonomous driving systems incorporate various sensor data representing different modalities of the same scene, which are interrelated with camera images. Incorporating such interrelation into existing image compression methods to enhance compression performance is feasible. Given that LiDAR point clouds represent a vital data modality, and their fusion with images has been shown to remarkably improve performance across a spectrum of semantic tasks\cite{liang2019multi, pang2020clocs, vora2020pointpainting, zheng2022boosting, krispel2020fuseseg, poliyapram2019point, zhang2019robust, chiu2021probabilistic}, we propose to incorporate sparse point cloud data to assist image compression.

In the field of image compression, traditional methods such as JPEG\cite{pennebaker1992jpeg}, JPEG-2000\cite{charrier1999jpeg2000}, BPG\cite{bpg} and VVC\cite{bross2021overview} have achieved decent performance. However, due to their predominantly handcrafted design, they lack scalability and struggle to incorporate information from other sensors. Recently, several learned image compression frameworks\cite{DBLP:conf/iclr/BalleLS17, DBLP:conf/iclr/BalleMSHJ18, DBLP:conf/nips/MinnenBT18, cheng2020learned, minnen2020channel, chen2021end, he2021checkerboard, he2022elic, zou2022devil, jiang2023mlic, ge2024nlic} have been proposed, demonstrating remarkably impressive performance. Some of these methods\cite{cheng2020learned, minnen2020channel, chen2021end, he2021checkerboard, he2022elic, zou2022devil, jiang2023mlic, ge2024nlic} are already comparable to the state-of-the-art traditional method VVC\cite{bross2021overview}, showcasing significant potential. In addition to surpassing traditional methods in performance, learned image compression, as a neural network-based approach, is inherently learnable. This makes it more flexible compared to traditional methods, allowing for easier integration of information from other sensors. However, the aforementioned methods solely utilize data from a single sensor and do not consider leveraging the correlation between multiple sensors to enhance performance.

To utilize information from multiple sensors, some researches\cite{liu2019dsic, huang2021l3c, deng2021deep, wodlinger2022sasic, lei2022deep, deng2023masic} employ learned image compression frameworks to compress stereo images. Stereo image compression aims to jointly compress pairs of stereoscopic images with left and right views. Leveraging the correlation between camera images from different perspectives leads to a significant performance improvement when jointly compressing two images compared to compressing a single image. However, despite the images being from different viewpoints, they belong to the same modality, making their relationship more closely connected compared to different modalities. Therefore, these methods cannot be directly applied to point cloud-assisted image compression.

In addition to single-modality approaches, other researches\cite{lu2022learning, lin2023your, gnutti2024lidar} leverage learned methods to utilize information from different modalities and enhance compression performance. Lu \textit{et al.}\cite{lu2022learning} use infrared images to assist in image compression, while Gnutti \textit{et al.}\cite{gnutti2024lidar} utilize point cloud depth maps captured by mobile phones for the same purpose. In comparison to these scenarios, LiDAR point clouds in autonomous driving are much sparser, contain less information, and have less correlation with camera images. These methods are not suitable for the scenario. Meanwhile, Lin \textit{et al.}\cite{lin2023your} use camera images to aid point cloud compression within the realm of autonomous driving, which is exactly the opposite of our research focus. However, it still employs a denser modality representation to enhance performance. The sparsity of LiDAR point clouds is the key distinction from the aforementioned scenarios, and it is also the primary challenge.

In this paper, we propose a new framework that uses sparse LiDAR point clouds to assist in learned image compression in the autonomous driving scenario. We first project the 3D sparse point cloud onto a 2D plane, resulting in a sparse depth map. Intending to extract structured information from the depth map, we design a series of point cloud feature extraction modules called Point-to-image Prediction (PIP) and Multi-scale Context Mining (MCM). PIP attempts to predict images from point cloud information, while MCM extracts multi-scale features from the predicted images. These features are further incorporated into existing learned image compression frameworks to enhance compression performance. Subsequently, we validate our approach across several learned compression networks, demonstrating its ability to improve compression efficiency. Visualization results further show our network's capability to extract denser structured information from sparse point clouds. In summary, our contributions are as follows:
\begin{itemize}
    \item We introduce a framework utilizing sparse point clouds to assist in image compression in the context of autonomous driving. To the best of our knowledge, this is the first method to use sparse point clouds to assist image compression.
    \item We design Point-to-image Prediction (PIP) and Multi-scale Context Mining (MCM) modules to extract dense structured information from sparse point clouds. Leveraging this information enhances compression performance and preserves more structural details in reconstructed images.
    \item We validate our approach across multiple existing learned image compression frameworks and achieve a noticeable improvement in compression performance.
\end{itemize}

The remainder of this paper is organized as follows. Sec. \ref{sec:related_works} provides an overview of the related literature. Sec. \ref{sec:method} details the model structure of the proposed sparse point cloud assisted learned image compression. Experimental results are discussed in Sec. \ref{sec:experiments}. Finally, Sec. \ref{sec:conclusion} concludes this paper.

\section{Related Works}
\label{sec:related_works}

In this section, we introduce the related work in literature. Considering that our work involves using multimodal information to assist image compression in autonomous driving scenarios, we categorize existing works into three main groups: learned lossy image compression, multi-sensor compression, and multi-modal tasks for autonomous driving. 

\subsection{Learned Lossy Image Compression}
Learned lossy image compression\cite{DBLP:conf/iclr/BalleLS17, DBLP:conf/iclr/BalleMSHJ18, DBLP:conf/nips/MinnenBT18, cheng2020learned, minnen2020channel, chen2021end, he2021checkerboard, he2022elic, zou2022devil, jiang2023mlic, zhang2023uniform, ge2024nlic, ma2019image} aims to establish a rate-distortion optimization (RDO) approach, aiming to compress images at the lowest possible bitrates $\mathcal{R}$ while maintaining a certain level of distortion $\mathcal{D}$. In a learning-based lossy image compression framework, it mainly consists of three parts: transformation, quantization, and entropy estimation.

Ball{\'{e}} \textit{et al.}\cite{DBLP:conf/iclr/BalleLS17} proposed adding uniform noise in place of actual quantization and introduced a method for bitrate estimation. Ball{\'{e}} \textit{et al.}\cite{DBLP:conf/iclr/BalleMSHJ18} also proposed a structure of hyperprior and introduced Gaussian distribution for probability estimation. Minnen \textit{et al.}\cite{DBLP:conf/nips/MinnenBT18} introduced a spatial autoregressive entropy model to extract contextual information in addition to hyperprior. He \textit{et al.}\cite{he2021checkerboard} further proposed a checkerboard pattern that takes advantage of contextual information while ensuring faster coding speed. In addition to the aforementioned work, some efforts have designed more effective entropy models for estimating the bitrate of $\bm{\hat{y}}$\cite{minnen2020channel, he2022elic, jiang2023mlic}, more complex transformation networks\cite{cheng2020learned, zou2022devil}, and considered optimizing the quantization operations\cite{zhang2023uniform, ge2024nlic}. While learning-based image compression methods have made significant progress in recent years, these methods are designed for single-modal single-image compression. However, with the proliferation of various modalities of data, compressing multiple types of data has become a topic worthy of exploration.

\begin{figure}[tbp]
  \centering
  \includegraphics[width=\linewidth]{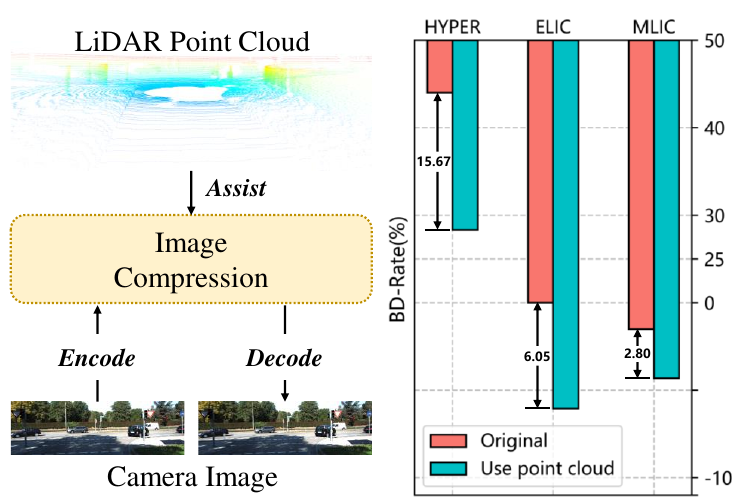}
  \caption{Image compression performance after using point clouds. Using sparse LiDAR point clouds to assist various image compression methods\cite{DBLP:conf/iclr/BalleMSHJ18, he2022elic, jiang2023mlic} can improve BD-Rate performance on the KITTI\cite{Geiger2013IJRR} dataset, with ELIC\cite{he2022elic} as the anchor.}
  \label{fig:mcm}
\end{figure}

\subsection{Multi-sensor Compression}
Multi-sensor compression involves combining data from multiple sensors to enhance overall compression performance. These sensor data can be of the same modality or different modalities. 

A typical task for utilizing multiple sensors with the same modality is stereo image compression, which aims to jointly compress pairs of images with different viewpoints. There have been many research works on traditional stereo image compression\cite{ellinas2004stereo, bezzine2018sparse, kadaikar2018joint}. Meanwhile, learning-based methods\cite{liu2019dsic, huang2021l3c, deng2021deep, wodlinger2022sasic, lei2022deep, deng2023masic} are also advancing rapidly. Liu \textit{et al.}\cite{liu2019dsic} proposed parametric skip functions and a conditional entropy model into deep stereo image compression. Deng \textit{et al.}\cite{deng2021deep} proposed a homography estimation based stereo image compression network. W{\"o}dlinger \textit{et al.}\cite{wodlinger2022sasic} proposed a lightweight network with latent shifts and stereo attention. Deng \textit{et al.}\cite{deng2023masic} further improved the network, achieving state-of-the-art results. Although these methods have made significant progress, the compressed data belongs to the same modality and cannot be applied to situations involving multiple modalities.

Compression between different modalities presents a richer landscape of scenarios. Hand-crafted methods focus on scenarios such as the combination of natural images with medical images\cite{brahimi2017improved}, or multi-view videos with depth maps\cite{varadarajan2012rgb, chen2020adaptive}. Learning-based methods include infrared images\cite{lu2022learning} or point clouds\cite{lin2023your, gnutti2024lidar}. While various modalities have been used for compression, due to the sparsity of point clouds, utilizing sparse LiDAR point clouds to assist in camera image compression is still a challenging task. Researches on this topic remain unexplored.

\begin{figure*}[tbp]
  \includegraphics[width=\textwidth]{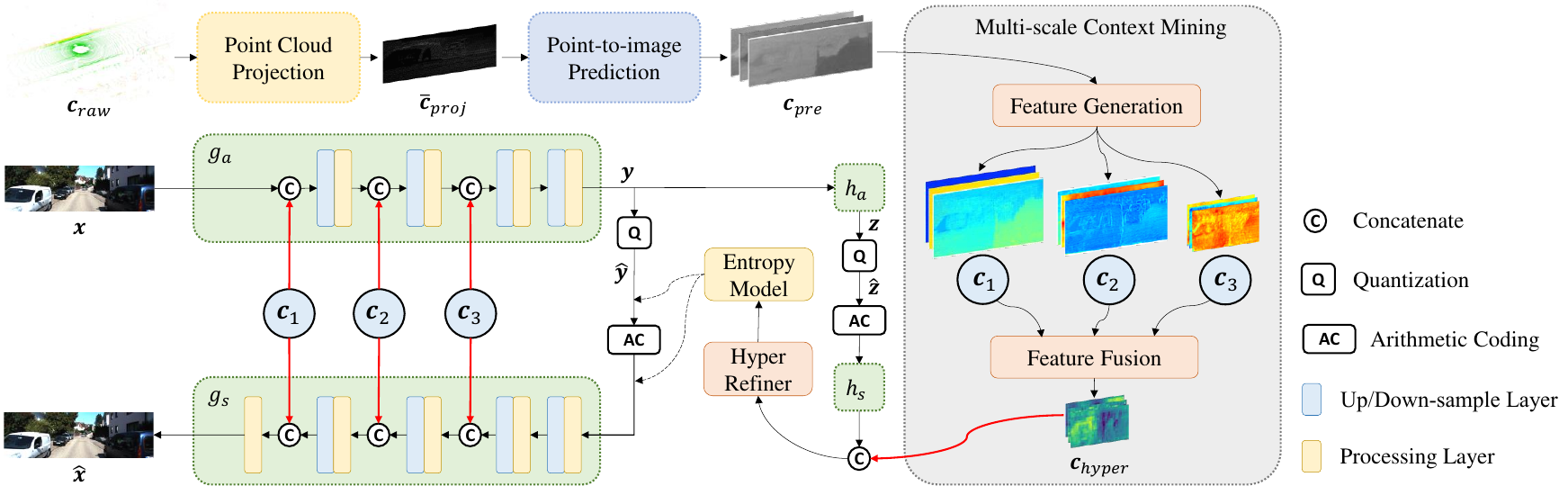}
  \caption{Overall architecture of our method. The blue layers in $g_{a}$ represent down-sampling, while in $g_{s}$, they represent up-sampling. The specific configuration of up/down-sample layers depends on the compression network used. The processing layers remain consistent with the original coding methods. For example, HYPER\cite{DBLP:conf/iclr/BalleMSHJ18} uses GDN/IGDNs\cite{DBLP:conf/iclr/BalleLS17} as processing layers, and ELIC\cite{he2022elic} uses Res-blocks\cite{he2016deep} or Attention-blocks\cite{vaswani2017attention}.}
  \label{fig:network_structure}
\end{figure*}

\begin{figure}[tbp]
  \centering
  \includegraphics[width=\linewidth]{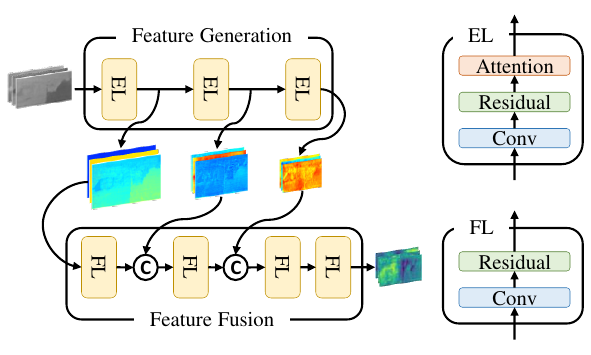}
  \caption{Detailed structure of Multi-scale Context Mining (MCM) module. The Extraction Layer (EL) consists of one convolutional layer, one res-block, and one attention-block. The Fusion Layer (FL) consists of one convolutional layer, and one res-block.}
  \label{fig:mcm}
\end{figure}

\subsection{Multi-modal Tasks for Autonomous Driving}
In the context of autonomous driving, the use of multi-modal information has been proven effective in many tasks. Some works\cite{liang2019multi, pang2020clocs, vora2020pointpainting, zheng2022boosting, song2024graph} demonstrate that combining LiDAR point clouds with camera images significantly improves performance in object detection. Other methods\cite{krispel2020fuseseg, poliyapram2019point, chang2023multiphase} focus on semantic segmentation, while some\cite{zhang2019robust, giefer2020eval, chiu2021probabilistic} utilize multi-modal information for object tracking.

In tasks leaning towards semantics, integrating multi-modal information is quite natural. These methods often exhibit significant advantages over single-modal approaches. However, in compression, effectively incorporating multi-modal information remains a challenging problem. Compared to semantic tasks, compression is a low-level task that is more difficult to leverage multi-modal information. Further research is needed in this area.

\begin{figure}[tbp]
  \centering
  \includegraphics[width=\linewidth]{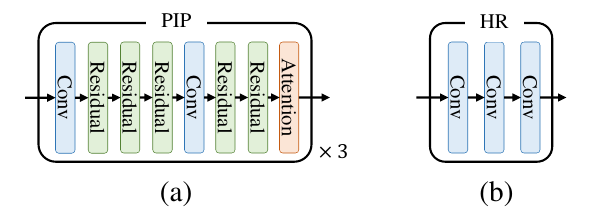}
  \caption{(a) Detailed structure of the Point-to-image Prediction (PIP) module. (b) Detailed structure of the Hyper Refiner (HR) module.}
  \label{fig:pip}
\end{figure}

\section{Method}
\label{sec:method}

In this section, we first introduce the overall framework of the proposed point cloud assisted image compression. Then, we introduce each step in order, including point cloud projection, point-to-image prediction, and multi-scale context mining. After that, we introduce some details of image compression used in the experiments.

\subsection{Overview}
Considering the scenario where point clouds and images representing the same scene coexist simultaneously, assuming priority transmission of the point cloud, our objective is to utilize the point cloud known at the encoding and decoding ends to assist in image compression. As shown in Figure ~\ref{fig:network_structure}, the overall architecture of our method has two parts: point cloud processing and image compression. 

For the point cloud processing branch, we first input the raw point cloud data into Point Cloud Projection (PCP) module, projecting the three-dimensional point cloud onto a two-dimensional plane. Then, the projected point cloud is input into Point-to-image Prediction (PIP) module to predict image information from the point cloud data. Subsequently, the predicted information is fed into Multi-scale Context Mining (MCM) module to extract multi-scale features, which are used in the image compression framework to enhance compression performance. 

For the image compression branch, we input the extracted multi-scale features from point cloud processing into both the transformation and entropy model of an existing image compression network. This process can be adapted to most typical learned image compression methods. Given a set of point cloud-related features $\bm{c} = \{\bm{c}_{1}, \bm{c}_{2}, \bm{c}_{3}, \bm{c}_{hyper} \}$ and an existing learned image compression network, the specific position where the point cloud features are introduced is illustrated in Figure ~\ref{fig:network_structure}. 

\subsection{Point Cloud Projection}
The purpose of point cloud projection is to project the 3D point cloud onto the perspective of a 2D image, aligning it spatially. To accurately project the point cloud onto a 2D plane, it generally requires several rotation and translation matrices. The specific procedure is as follows:
\begin{equation}
    \bm{T}_{LiD}^{cam} = 
        \begin{pmatrix}
            \bm{R}_{LiD}^{cam} & \bm{t}_{LiD}^{cam} \\
            0 & 1
        \end{pmatrix},
\end{equation}
\begin{equation}
    \bm{c}_{proj} = \bm{P}_{rect}\bm{R}_{rect}\bm{T}_{LiD}^{cam}\bm{c}_{raw},
\end{equation}
where $\bm{c}_{raw} \in \mathbb{R}^{N \times 4}$ represents the raw 3D point cloud and $\bm{c}_{proj} \in \mathbb{R}^{1 \times H \times W}$ represents the projected point cloud. $N$ is the number of points in the point cloud, and $H, W$ represent the height and width of the output image, respectively. $\bm{R}_{LiD}^{cam}$ and $\bm{t}_{LiD}^{cam}$ represent the rotation matrix and translation vector from LiDAR to camera. $\bm{R}_{rect}$ is the rectifying rotation matrix. $\bm{P}_{rect}$ is the projection matrix after rectification.

\begin{figure}[tbp]
  \centering
  \includegraphics[width=\linewidth]{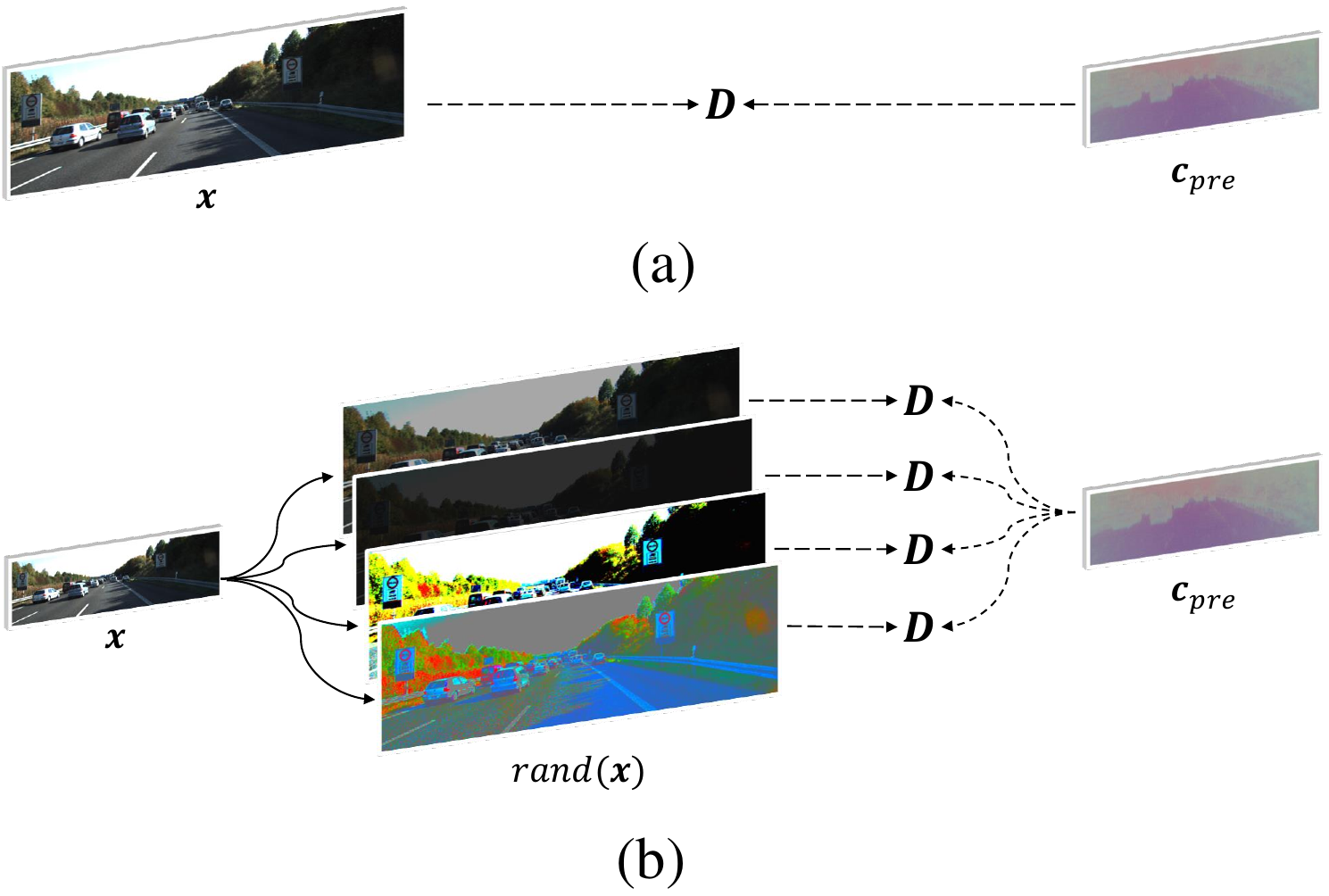}
  \caption{(a) Directly predicting RGB values using point cloud is very challenging. (b) Predicting color-transformed images can learn a fine-grained structural feature.}
  \label{fig:pip_concept}
\end{figure}

After the aforementioned process, the values of elements in $\bm{c}_{proj}$ represent the depth of points in 3D space from the viewing plane, effectively forming a depth map. However, depth and RGB values are not directly related; instead, they share consistency in the structure of objects. Since we only aim to utilize this structural similarity to assist in image compression, we do not need highly accurate depth information. In fact, since the depth range varies significantly across different scenes, having overly precise depth information would increase the difficulty for the model to process. Therefore, we performed normalization and histogram equalization on the depth map to a fixed range, ensuring that only the structural information remains consistent. The specific operations are as follows:

\begin{equation}
    \bm{\overline{c}}_{proj} = Hist(\lceil (\bm{c}_{proj} - min(\bm{c}_{proj})) \times s \rfloor),
\end{equation}
where $\bm{\overline{c}}_{proj} \in \mathbb{R}^{1 \times H \times W}$ represents the depth map after normalization and histogram equalization. $s$ is a pre-defined hyperparameter used to represent the scaling range. $Hist(\cdot)$ is the histogram equalization operation.

\subsection{Point-to-image Prediction}
Although the point cloud is projected into a depth map, aligning spatially with the input image, it is much sparser compared to images. Directly using a sparse depth map to assist dense image compression is highly challenging.

To narrow the gap between these two modalities, we employ the Point-to-image Prediction (PIP) module, constructed by connecting multiple convolutional layers, Res-blocks\cite{he2016deep}, and Attention-blocks\cite{vaswani2017attention} as shown in Figure~\ref{fig:pip}(a). The purpose of PIP is to predict denser structural information directly from the sparse depth map, aligning the depth map to the corresponding image. This process can be formulated as:
\begin{equation}
    \bm{c}_{pre} = PIP(\bm{\overline{c}}_{proj}),
\end{equation}
where $\bm{c}_{pre} \in \mathbb{R}^{3 \times H \times W}$ represents the predicted feature. $PIP(\cdot)$ represents the neural network-based point-to-image prediction module. 

\begin{figure}[tbp]
  \centering
  \includegraphics[width=\linewidth]{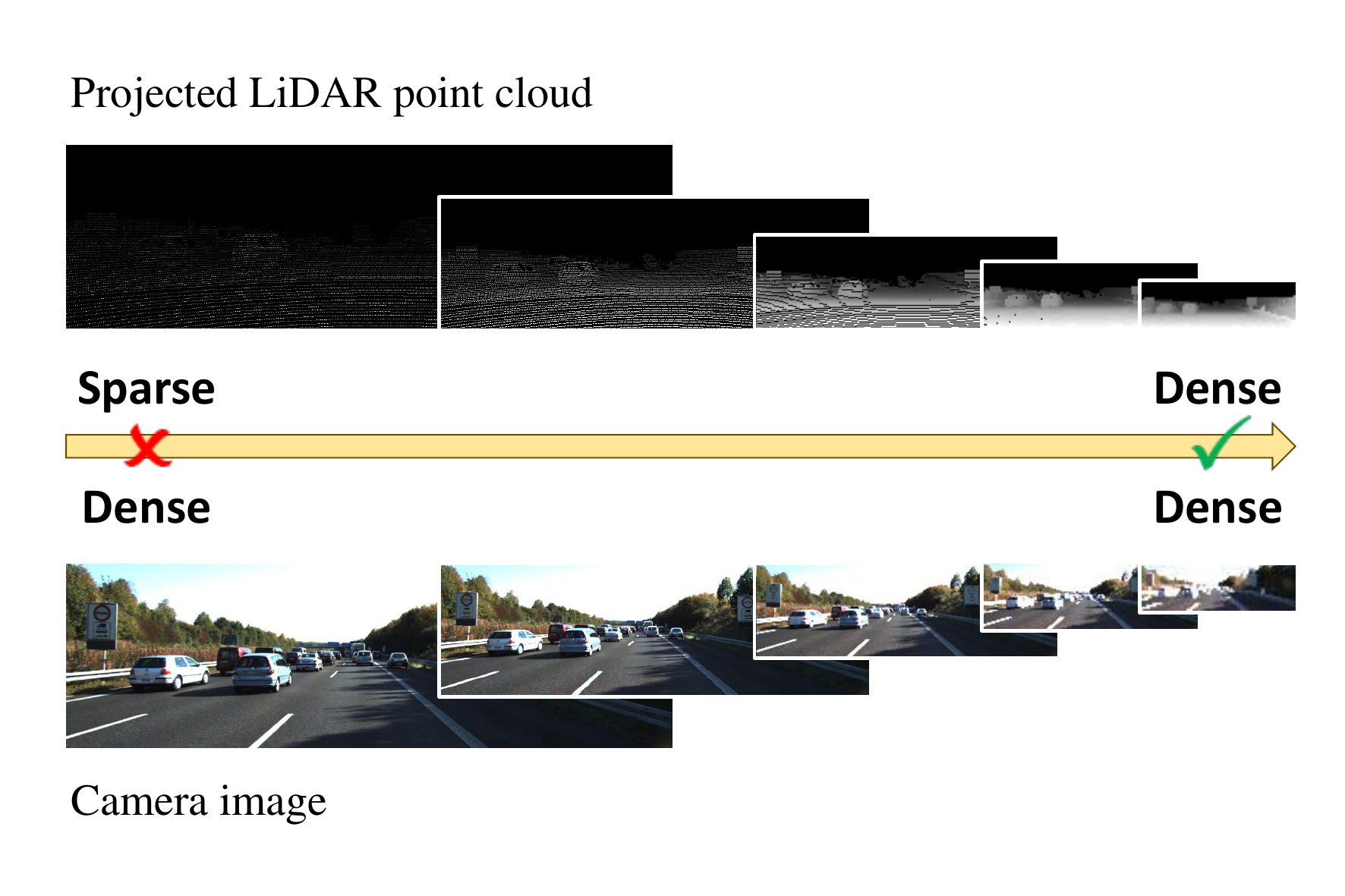}
  \caption{Sparse point clouds become dense after several down-sampling layers.}
  \label{fig:mcm_concept}
\end{figure}

 Since image compression operates at the pixel level, $\bm{c}_{pre}$ needs to provide structural information with a granularity similar to that of camera images to better assist in this task. A very natural idea is to directly predict the RGB values, as the RGB image itself meets the requirements in terms of proper structural information and fine-grained density. However, depth maps have no direct relationship with RGB values, making it difficult to predict accurate RGB values directly from depth maps. Therefore, we randomly alter the RGB values, which makes the training target changing within a certain range each time. This transforms the task from "predicting accurate RGB values" to "predicting an RGB range," thereby reducing the original task's difficulty. These operations include adjusting contrast, brightness, and inverting colors. As shown in Figure~\ref{fig:pip_concept}, random operations ensure that the RGB values of each prediction target are different each time, while the fine-grained image structure remains the same. The loss function for this step is formalized as:
\begin{equation}
    \mathcal{L}_{pre} = \mathcal{D}(rand(\bm{x}), \bm{c}_{pre}),
\end{equation}
where $\bm{x} \in \mathbb{R}^{3 \times H \times W}$ represents the input image. $rand(\cdot)$ represents the random color transformation operations.

\subsection{Multi-scale Context Mining}

Although the PIP module generates denser structural features compared to point clouds, it is still insufficient for image compression. Since point cloud information is much sparser than images, the capability of a single-scale feature generated by a single PIP module is limited. To better leverage the structural information provided by the point cloud, we need to perform further feature extraction on the prediction results. Inspired by the learned video compression\cite{sheng2022temporal}, we consider introducing a Multi-scale Context Mining (MCM) module as shown in Figure~\ref{fig:mcm}, which further extracts features of different scales from $\bm{c}_{pre}$. As shown in Figure~\ref{fig:mcm_concept}, point clouds cannot be matched with images at the initial scale due to sparsity. After multiple rounds of down-sampling, the sparse point cloud becomes dense, allowing it to be matched with images of the same scale. However, if the scale is too small, too much information will be lost. Multi-scale operations enable the matching of point clouds and images at various scales, maximizing the utilization of point cloud information while avoiding excessive information loss.

The MCM module consists of two parts. The first part is multi-scale feature generation (FG), which generates three different scales of point cloud features. It is formalized as follows:
\begin{equation}
    \bm{c}_{1}, \bm{c}_{2}, \bm{c}_{3} = FG(\bm{c}_{pre}),
\end{equation}
where $\bm{c}_{1} \in \mathbb{R}^{C \times H \times W}, \bm{c}_{2} \in \mathbb{R}^{C \times \frac{H}{2} \times \frac{W}{2}}, \bm{c}_{3} \in \mathbb{R}^{C \times \frac{H}{4} \times \frac{W}{4}}$ represent the point cloud features at three different scales, with $C$ being the pre-defined number of channels. $FG(\cdot)$ represents the network-based multi-scale feature generation.

The second part is multi-feature fusion (FF), which merges the three features into one. It is formalized as follows:
\begin{equation}
    \bm{c}_{hyper} = FF(\bm{c}_{1}, \bm{c}_{2}, \bm{c}_{3}),
\end{equation}
where $\bm{c}_{hyper} \in \mathbb{R}^{C^{'} \times \frac{H}{16} \times \frac{W}{16}}$ represents the fused feature, with $C^{'}$ being the pre-defined number of channels. $FF(\cdot)$ represents the network-based multi-feature fusion.

After the MCM module, we obtain all the point cloud features for image compression:
\begin{equation}
    \bm{c} = \{\bm{c}_{1}, \bm{c}_{2}, \bm{c}_{3}, \bm{c}_{hyper} \}.
\end{equation}

\subsection{Image Compression}

The extracted point cloud features $\bm{c}$ will be utilized across different parts of the learned image compression network, which is a multi-modal compression framework designed based on existing methods \cite{DBLP:conf/iclr/BalleMSHJ18, he2022elic, jiang2023mlic}.

The network consists of an analyzer $g_{a}$, a synthesizer $g_{s}$, a hyper analyzer $h_{a}$, a hyper synthesizer $h_{s}$, and an entropy model, where the entropy model, $h_{a}$, and $h_{s}$ are the same as those in the original compression model. The remaining parts have undergone certain modifications to incorporate the point cloud information as shown in Figure~\ref{fig:network_structure}. In both the encoder and decoder, $\bm{c}_{1}, \bm{c}_{2}, \bm{c}_{3}$ are concatenated with image features of the same scale, and then sent to the next layer of the network. Meanwhile, we add a Hyper Refiner (HR) module to fuse the hyper information of point clouds and images. The HR module consists of three convolutional layers as shown in Figure~\ref{fig:pip}(b). Specific operations is formalized as follows:
\begin{equation}
    \bm{\hat{y}} = \lceil \bm{y} \rfloor = \lceil g_{a}(\bm{x}|\bm{c}_{1}, \bm{c}_{2}, \bm{c}_{3}) \rfloor,
\end{equation}
\begin{equation}
    \bm{\hat{x}} = g_{s}(\bm{\hat{y}}|\bm{c}_{1}, \bm{c}_{2}, \bm{c}_{3}),
\end{equation}
where $\bm{\hat{y}}$ represents the conditional cross-modal discrete latent to be compressed. $\bm{x} \in \mathbb{R}^{3 \times H \times W}$ and $\bm{\hat{x}} \in \mathbb{R}^{3 \times H \times W}$ are the input image and the reconstructed image, respectively. For the hyper part, we can formalize as follows:
\begin{equation}
    \bm{\hat{z}} = \lceil h_{a}(\bm{y}) \rfloor,
\end{equation}
\begin{equation}
    \bm{z}_{hyper} = h_{s}(\bm{\hat{z}}),
\end{equation}
\begin{equation}
    \bm{\mu}_{hyper}, \bm{\sigma}_{hyper} = HR(\bm{z}_{hyper}, \bm{c}_{hyper}),
\end{equation}
where $\lceil \cdot \rfloor$ represents the quantization operation. $HR(\cdot)$ represents the Hyper Refiner module. $\bm{\mu}_{hyper}$ and $\bm{\sigma}_{hyper}$ will replace the corresponding parts of the original compression model as the input for the entropy model, used to estimate the bitrate of $\bm{\hat{y}}$.

In summary, the final loss function is formalized as:
\begin{equation}
    \mathcal{L} = \mathcal{R}(\bm{\hat{y}}|\bm{\hat{z}}, \bm{c}) + \mathcal{R}(\bm{\hat{z}}) + \lambda \mathcal{D}(\bm{x}, \bm{\hat{x}}) + \alpha \mathcal{L}_{pre},
\end{equation}
where $\mathcal{R}$ and $\mathcal{D}$ represent bitrates and distortion, respectively. $\lambda$ and $\alpha$ are pre-defined hyperparameters. $\alpha$ is used to control the proportion of $\mathcal{L}_{pre}$ in the total loss function. In our design, $\alpha$ gradually decreases with the training epochs.

\section{Experiments}
\label{sec:experiments}

In this section, we present some details of our experiments, including experiment settings, compression performance, ablation experiments, and complexity.

\subsection{Experiment Settings}
We select KITTI\cite{Geiger2013IJRR} and Waymo\cite{sun2020scalability} as our datasets, which contain image frames and point cloud frames captured at the same time. The point cloud frames are assumed to have been losslessly compressed. The detailed configurations are as follows:

\begin{figure}[tbp]
  \centering
  \includegraphics[width=1\linewidth]{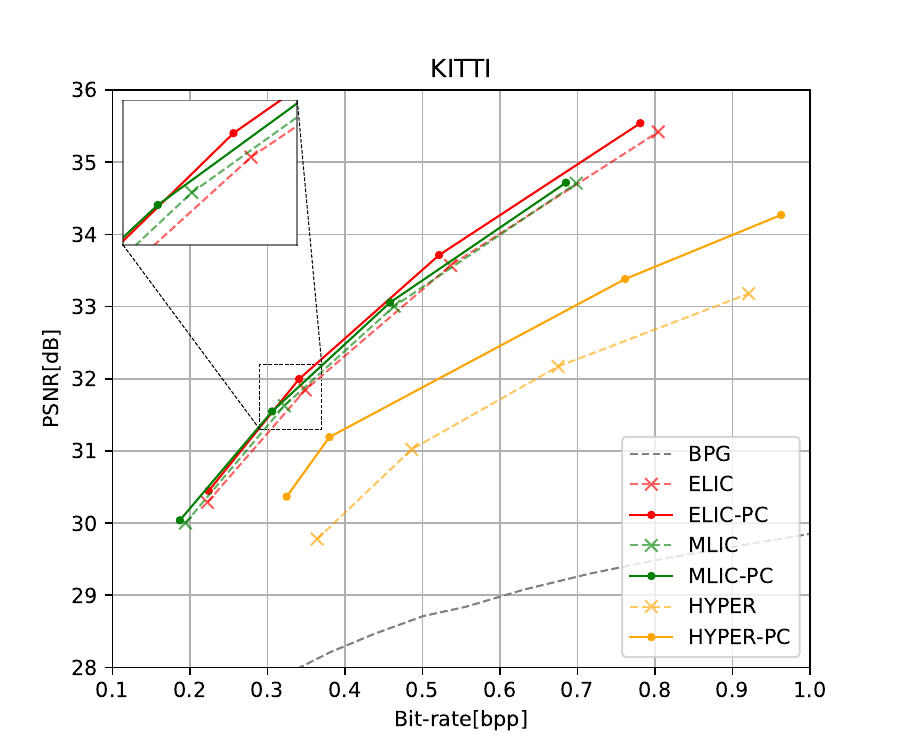}
  \caption{RD performance of different methods on KITTI. ``PC" refers to incorporating point cloud information into the model. BPG\cite{bpg} is a well-known traditional compression method.}
  \label{fig:main}
\end{figure}

\begin{table}
\caption{BD-Rate (\%) for PSNR (dB) of different models. The anchor is ELIC\cite{he2022elic}. ``$\triangle$" represents the performance improvement of using point cloud information compared to the original one. ``$\downarrow$" represents that the lower the metric, the better.}
  \begin{tabular}{lccccc}
    \toprule
    \multirow{2}{*}{Model} & \multicolumn{2}{c}{KITTI}                          & \multicolumn{2}{c}{Waymo}                          \\
                           & BD-Rate(\%) $\downarrow$  & $\triangle \downarrow$ & BD-Rate(\%) $\downarrow$  & $\triangle \downarrow$ \\
    \midrule
     ELIC\cite{he2022elic} & 0.00 & - & 0.00 & -\\
     ELIC-PC & -6.05 & -6.05 & -5.99 & -5.99\\
     HYPER\cite{DBLP:conf/iclr/BalleMSHJ18} & +43.98 & - & +53.88 & -\\
     HYPER-PC & +28.31 & -15.67 & +36.71 & -17.17 \\
     MLIC\cite{jiang2023mlic} & -1.51 & - & -2.92 & -\\
     MLIC-PC & -4.31 & -2.80 & -7.97 & -5.05\\
  \bottomrule
\end{tabular}
\label{tab:main}
\end{table}

\subsubsection{KITTI Dataset}
We select 12,674 frames from the raw data for training, 1,534 frames for validation, and 3,114 frames for testing. Training, validation, and testing datasets are sampled from different scenes. Since the projected point cloud covering all the objects of interest on the image, with the uncovered areas being the distant scenery which are not of our concern, we crop the camera images to a resolution of $1242 \times 256$ pixels to better verify the effectiveness of the algorithm. Point Cloud Projection (PCP) is executed first and obtains $\bm{\overline{c}}_{proj}$, which is stored as an array.
\subsubsection{Waymo Dataset}
We select 13,800 frames from the raw data for training, 2,300 frames for validation, and 2,300 frames for testing. For the same reason as KITTI, we crop the images to a resolution of $1920 \times 760$ pixels. The remaining settings are the same as those in KITTI.
\subsubsection{Training Details}
During training, we randomly crop images as well as the stored $\bm{\overline{c}}_{proj}$ to $256 \times 256$ patches. For each model, we use different $\lambda$ to control the bitrate range. We set $\lambda \in \{4, 8, 16, 32\} \times 10^{-3}$ when optimizing MSE. We train each model with an Adam optimizer with $\beta_{1} = 0.9$, $\beta_{2} = 0.999$. We train our models for 1M steps with a batch size set to 8 and a learning rate set to $10^{-4}$. We set $\alpha = 0.01$ for the first 500K steps, $\alpha = 0.005$ between 500-900K steps, and $\alpha = 0$ for the remaining part.
\subsubsection{Compression Models}
We select three baseline models, namely HYPER\cite{DBLP:conf/iclr/BalleMSHJ18}, ELIC\cite{he2022elic}, and MLIC\cite{jiang2023mlic}, to which we add the point cloud branch to validate our approach. The specific architectures of these three models remain consistent with the settings described in their respective papers.

\begin{figure}[tbp]
  \centering
  \includegraphics[width=1\linewidth]{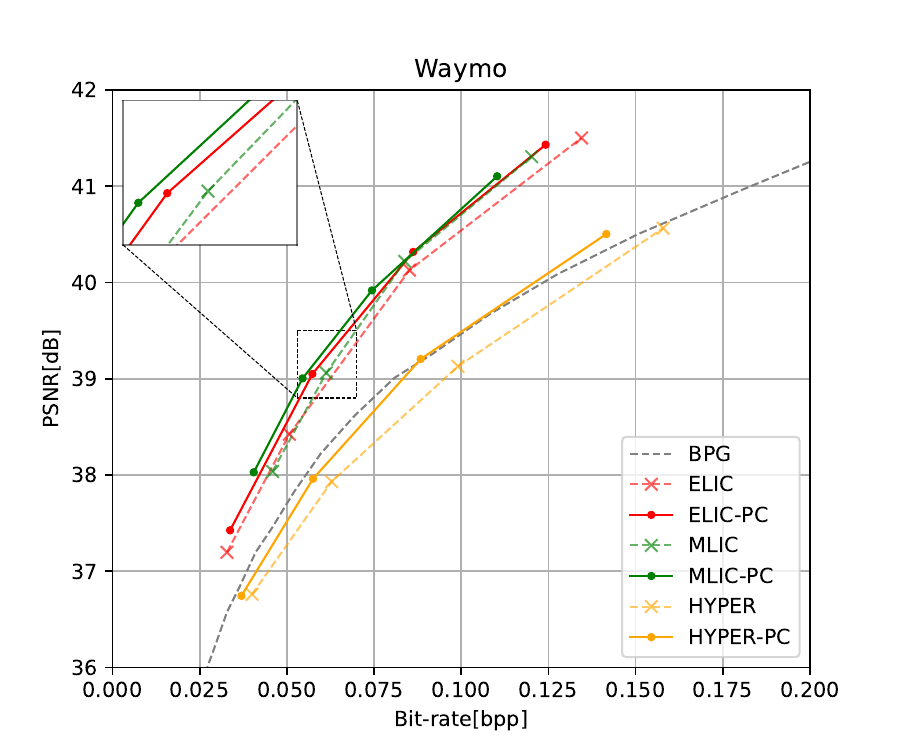}
  \caption{RD performance of different methods on Waymo. ``PC" refers to incorporating point cloud information into the model. BPG\cite{bpg} is a well-known traditional compression method.}
  \label{fig:main_waymo}
\end{figure}

\subsection{Performance}
We evaluate the compression performance of various models with and without our point cloud assistance framework and conduct a visual analysis of some features of our method.

\subsubsection{Rate-distortion Performance}
As shown in Figure~\ref{fig:main} and Table~\ref{tab:main}, we test the original compression performance of three models as well as their performance after incorporating point cloud information on the KITTI dataset. Following the same training strategy, the original model performance ranges from worst to best as HYPER, ELIC, and MLIC, with the models' designs becoming more complex as performance improved. Using ELIC as the anchor, after incorporating point cloud information, HYPER improves by 15.67\%, ELIC improves by 6.05\%, and MLIC improves by 2.80\% on KITTI. Compared to KITTI, the Waymo dataset has a larger scale and encompasses a richer variety of scenes. Importantly, the environment in Waymo is more complex, including scenarios such as night, rain, fog, and other scenes that cannot be processed by point cloud information. Therefore, experiments on Waymo present a greater challenge. As shown in Figure~\ref{fig:main_waymo} and Table~\ref{tab:main}, with ELIC as the baseline, after incorporating point cloud information, HYPER improves by 17.17\%, ELIC improves by 5.99\%, and MLIC improves by 5.05\% on Waymo. The above results indicate that different models experience performance improvements when incorporating a point cloud branch, with simpler models showing larger improvements.

\begin{figure*}[tbp]
  \includegraphics[width=\textwidth]{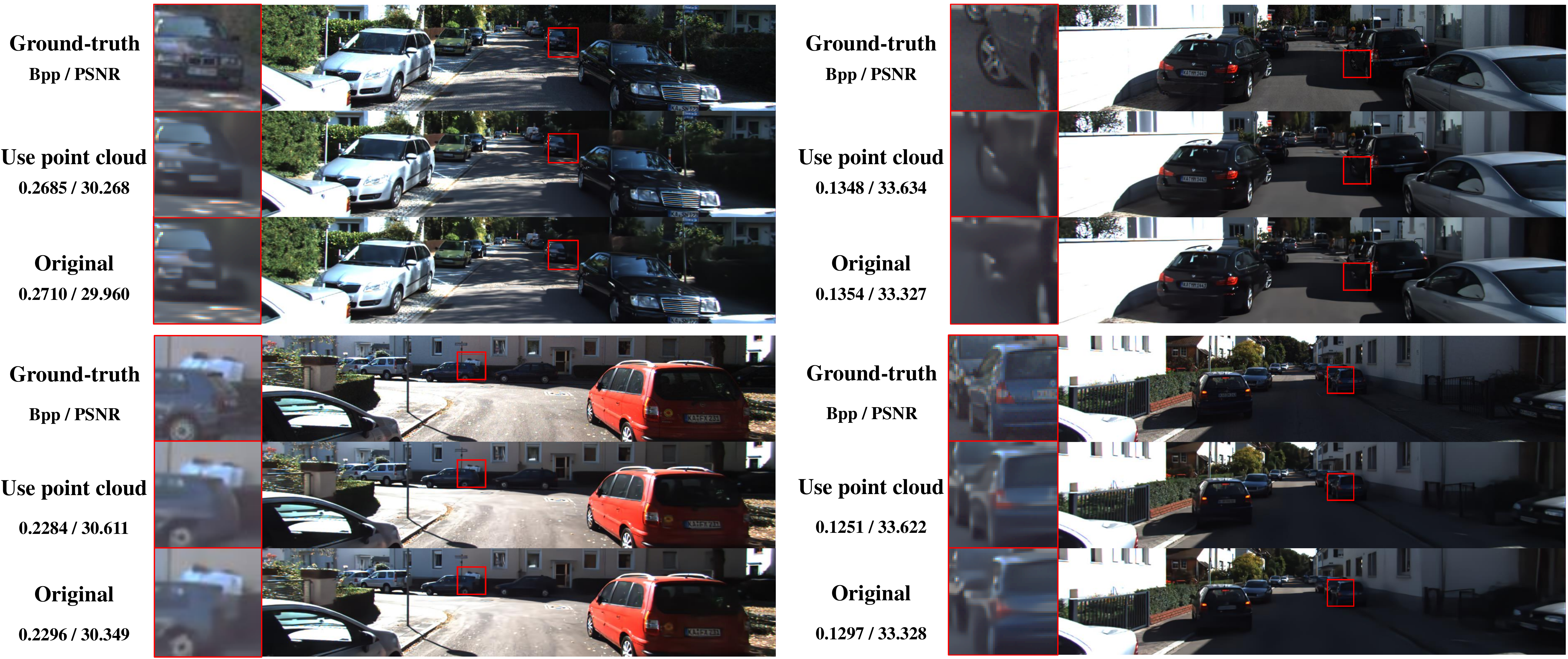}
  \caption{Visualization of reconstructed images on KITTI. The metric is [Bpp$\downarrow$ / PSNR$\uparrow$]. We compared the results of ELIC\cite{he2022elic} ($\lambda = 0.004$) using the original method with those after incorporating point cloud information.}
  \label{fig:visual_pic}
\end{figure*}

\begin{figure}[tbp]
  \centering
  \includegraphics[width=1\linewidth]{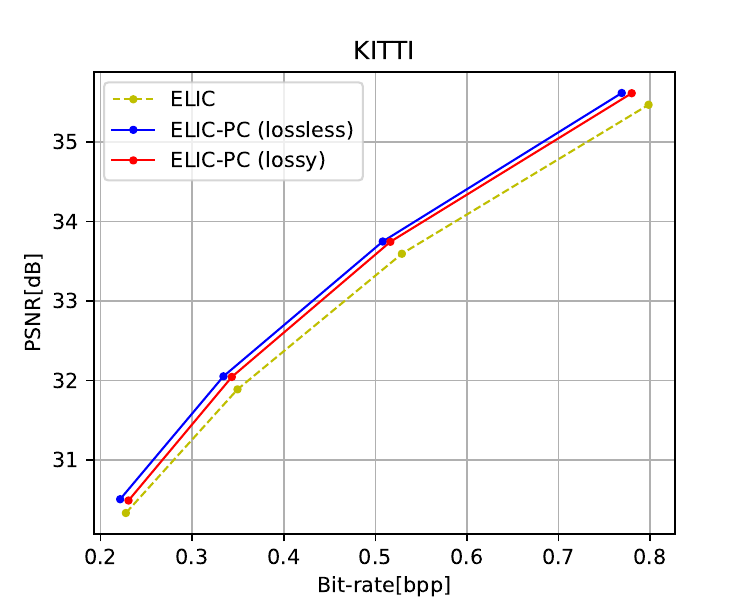}
  \caption{RD performance of ELIC-PC when point clouds are lossy compressed. The model is not retrained.}
  \label{fig:compressed}
\end{figure}

\subsubsection{Performance of Lossy Compressed Point Clouds}
To verify the robustness of our method, we subject the point cloud to lossy compression and examine whether the model could still improve compression performance with lossy point cloud inputs. The point cloud compression method we utilize is G-PCC\cite{gpcc}, and we choose the configuration with the highest officially provided bitrate for lossy compression. Instead of retraining the model, we directly employ the network trained on lossless point clouds to conduct this test. In comparison to the lossless scenario, using lossy compressed point clouds results in a degradation of 2.41\% in RD performance on the KITTI dataset as shown in Figure~\ref{fig:compressed}, yet it still outperforms scenarios where point clouds are not used. It is worth noting that the loss at low bitrates is higher than at high bitrates, indicating that the accuracy of point cloud information is more crucial for low bitrates.

\subsubsection{Visualization}
\begin{figure*}[tbp]
  \includegraphics[width=\textwidth]{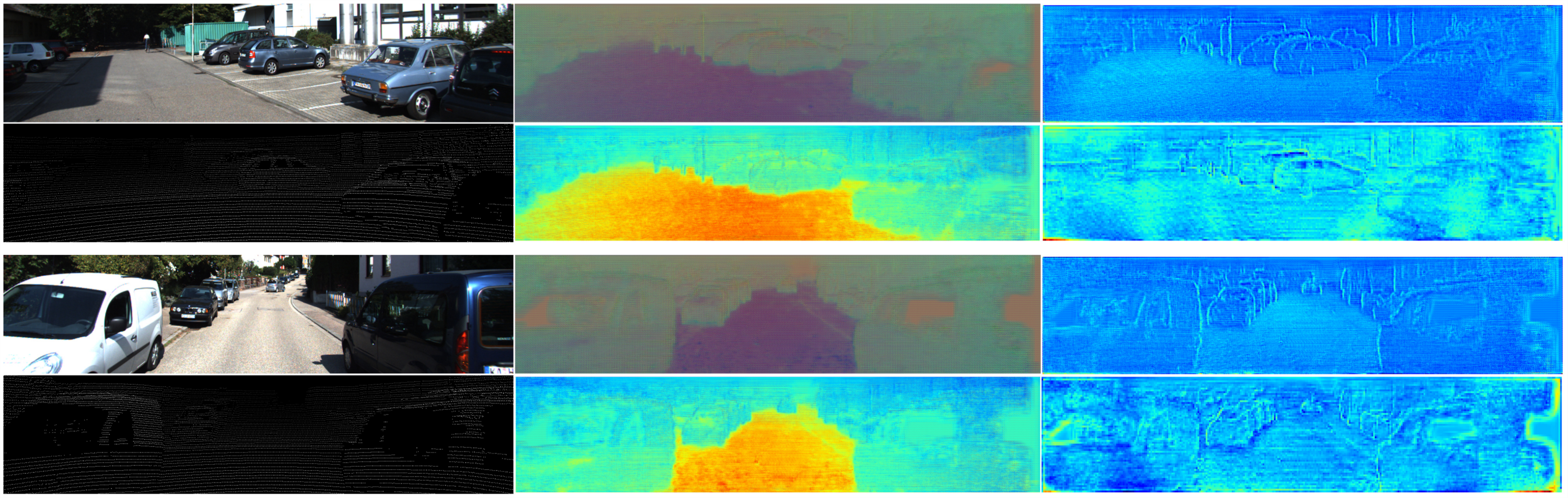}
  \caption{Visualization of the features from the point cloud branch. From top to bottom, and from left to right, respectively, we have the original image $\bm{x}$, the projected point cloud $\bm{\overline{c}}_{proj}$, the predicted image $\bm{c}_{pre}$, and the features $\bm{c}_{1}, \bm{c}_{2}, \bm{c}_{3}$ at three scales. Use ELIC-PC model with $\lambda = 0.032$.}
  \label{fig:visual}
\end{figure*}

We conduct a visual analysis of the reconstructed images and the features extracted by the point cloud branch of our method. From the reconstructed images in Figure~\ref{fig:visual_pic}, without using point cloud information, the original model exhibits distortion on some smaller object structures. However, when incorporating point cloud information, even small structures can be preserved intact, indicating that our method not only exhibits better compression performance but also more accurately captures structures in some detail areas. From Figure~\ref{fig:visual}, the projected point cloud $\bm{\overline{c}}_{proj}$ is highly sparse compared to the camera image $\bm{x}$, making it challenging to discern the original image structure. However, after passing through the PIP module, the feature $\bm{c}_{pre}$ becomes denser and can more clearly distinguish certain structures, such as the ground and vehicles. Furthermore, after the MCM module, the structural features $\bm{c}_{1}, \bm{c}_{2}, \bm{c}_{3}$ become even clearer, especially in the second scale feature $\bm{c}_{2}$, which closely resembles the original image structure. Based on these visualizations, it is evident that our method is capable of extracting structural features highly similar to the original image from sparse point clouds.

\subsection{Ablation Studies}
We conduct ablation studies using the ELIC model on KITTI, analyzing the performance of each module proposed. All training settings remain consistent with the previous setup.

\subsubsection{Analysis of Point-to-image Prediction Module}
\begin{figure}[tbp]
  \centering
  \includegraphics[width=1\linewidth]{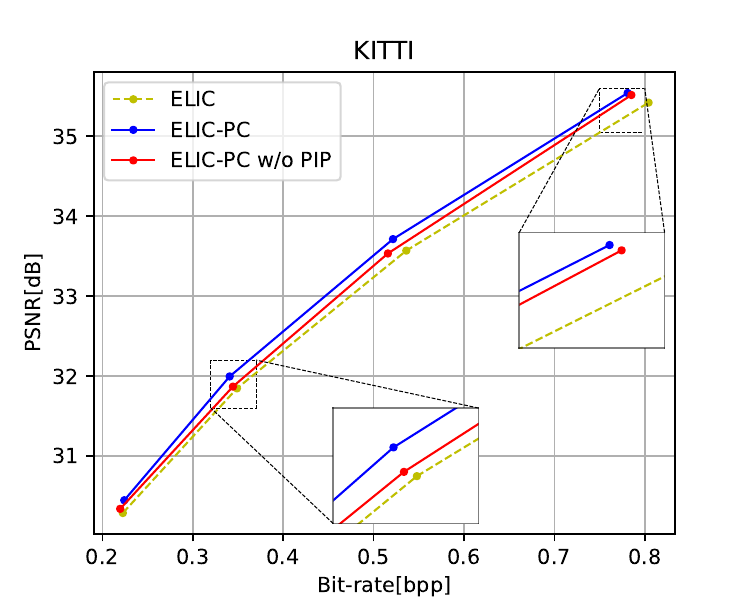}
  \caption{RD performance of PIP module ablation experiment.}
  \label{fig:pip_study}
\end{figure}

We remove the PIP module and directly input $\bm{\overline{c}}_{proj}$ into the MCM module to extract features for analyzing the impact of the PIP module on compression performance. As shown in Table~\ref{tab:mcm_study}, the overall compression performance decreases after removing the PIP module, with a performance loss of 3.16\%. It is worth noting that the PIP module has a greater impact on the low-bitrate end and a smaller impact on the high-bitrate end as shown in Figure~\ref{fig:pip_study}. It is possibly because low-bitrate models tend to rely more on point cloud information to assist image compression, while high-bitrate ends do not require as much assistance from point clouds due to sufficient information capacity.

\begin{table}
\centering
\caption{Ablation studies of PIP and MCM on KITTI. The metric is BD-Rate (\%) for PSNR (dB). The anchor is ELIC\cite{he2022elic}. ``$\downarrow$" represents that the lower the metric, the better.}
  \begin{tabular}{lccccc}
    \toprule
    Model & PIP & FG & FF & BD-Rate(\%) $\downarrow$\\
    \midrule
     ELIC & & & & 0.00\\
     ELIC-PC w/o PIP &  & $\bm{\checkmark}$ & $\bm{\checkmark}$ & -2.89 \\
     ELIC-PC w/o FG & $\bm{\checkmark}$ & & $\bm{\checkmark}$ & -0.88\\
    ELIC-PC w/o FF & $\bm{\checkmark}$ & $\bm{\checkmark}$ & & -4.13\\
    ELIC-PC & $\bm{\checkmark}$ & $\bm{\checkmark}$ & $\bm{\checkmark}$ & -6.05\\
  \bottomrule
\end{tabular}
\label{tab:mcm_study}
\end{table}

\begin{figure}[tbp]
  \centering
  \includegraphics[width=1\linewidth]{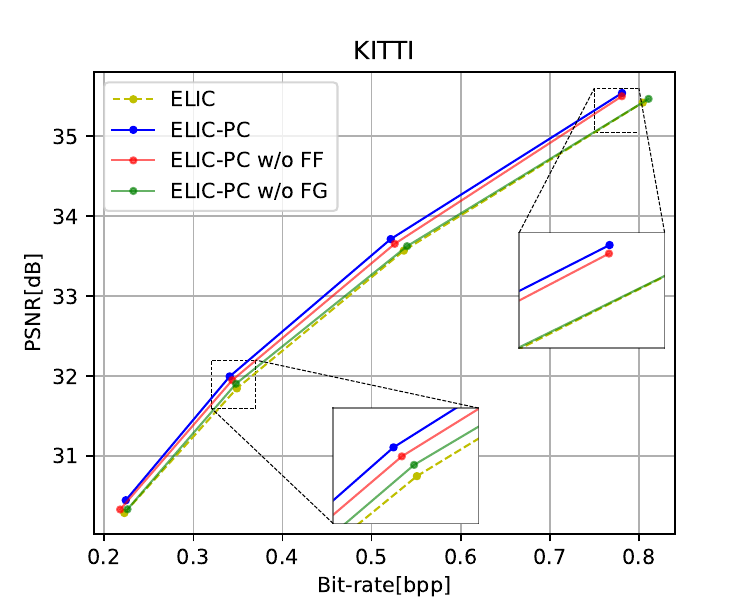}
  \caption{RD performance of MCM module ablation experiment.``FG" stands for Feature Generation, and ``FF" stands for Feature Fusion.}
  \label{fig:mcm_study}
\end{figure}

\subsubsection{Analysis of Multi-scale Context Mining Module}
We analyze the performance of the MCM module. As mentioned earlier, the MCM module consists of two parts: Feature Generation (FG) and Feature Fusion (FF). We use exactly the same model structure, with the difference being that the generated features are not incorporated into the corresponding part of the compression model. For ``ELIC-PC w/o FG", we do not input $\bm{c}_{1}, \bm{c}_{2}, \bm{c}_{3}$ into transform; for ``ELIC-PC w/o FF", we do not input $\bm{c}_{hyper}$ into the Hyper Refiner. The experimental results are shown in Figure~\ref{fig:mcm_study} and Table~\ref{tab:mcm_study}. Both FF and FG modules enhance the model's performance to some extent. Compared to not using point cloud information, using only the FF module to enhance the Hyper Refiner yields a performance gain of 0.88\%, while using only the FG module to enhance transform results in a performance gain of 4.13\%. Surprisingly, simultaneously using both FF and FG modules provides a gain of 6.05\%, which exceeds the sum of the gains from using each module separately. This indicates that the gains brought by these two modules do not overlap, instead, combining them further enhances performance. From Figure~\ref{fig:mcm_study}, it can be inferred that FG plays a major role in the high-bitrate end, with almost no performance loss if FF is not utilized. If only FF is employed, the performance will drop to a level similar to that of the original ELIC. However, at lower bitrate ends, both FG and FF make certain contributions. Nevertheless, the contribution of FG is significantly higher than that of FF.

\begin{figure}[tbp]
  \centering
  \includegraphics[width=1\linewidth]{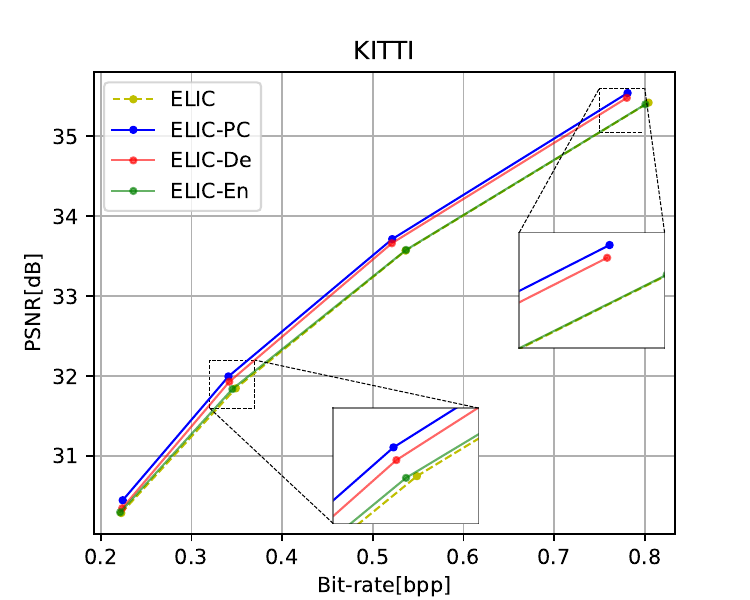}
  \caption{RD performance using point cloud information only at the encoding or decoding side. ``En" indicates using point clouds only on the encoding side, while ``De" indicates using it only on the decoding side.}
  \label{fig:en_de}
\end{figure}

\begin{table}
\centering
\caption{Ablation studies of coding sides and parameter quantity on KITTI. The metric is BD-Rate (\%) for PSNR (dB). The anchor is ELIC\cite{he2022elic}. ``$\downarrow$" represents that the lower the metric, the better.}
  \begin{tabular}{lccccc}
    \toprule
    \multirow{2}{*}{Model} & \multirow{2}{*}{Encoder} & \multirow{2}{*}{Decoder} & Parameter    & \multirow{2}{*}{BD-Rate(\%) $\downarrow$} \\
                           &                          &                          & Augmentation &                                           \\
    \midrule
     ELIC & & & & 0.00\\
     ELIC(PA) & & & $\bm{\checkmark}$ & -3.24\\
     ELIC-En & $\bm{\checkmark}$ & & & -0.42\\
    ELIC-De & & $\bm{\checkmark}$ & & -4.32\\
    ELIC-PC & $\bm{\checkmark}$ & $\bm{\checkmark}$ & & -6.05 \\
  \bottomrule
\end{tabular}
\label{tab:code}
\end{table}

\subsubsection{Decorrelation versus Reconstruction}
To further investigate how point cloud information assists in image compression, we consider it from both encoding and decoding perspectives. If point cloud information contributes more to compression during encoding, it indicates that it helps the image be better decorrelated. If it contributes more during decoding, it suggests that the point cloud information aids in better image reconstruction. As shown in Figure~\ref{fig:en_de} and Table~\ref{tab:code}, introducing point cloud information only at the encoding side yields a performance gain of 0.42\%, while introducing it only at the decoding side results in a performance gain of 4.32\%. This indicates that utilizing point cloud information during decoding leads to greater performance enhancement. Since the information provided remains identical at both encoding and decoding ends, it can be inferred that point cloud information tends to optimize image reconstruction.

\subsection{Complexity}
We analyze the complexity of our method, including the impact of parameter quantity, the impact of model structure, parameter size, and coding time.

\subsubsection{Impact of Parameter Quantity}
\begin{figure}[tbp]
  \centering
  \includegraphics[width=\linewidth]{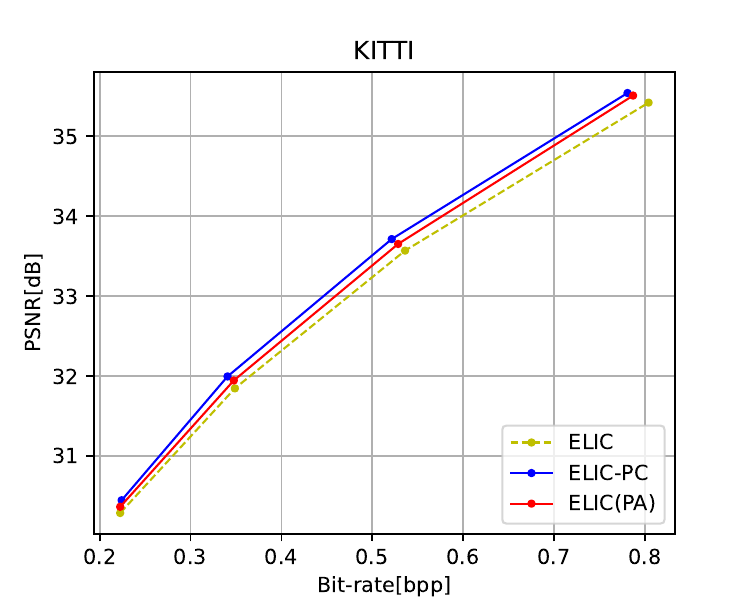}
  \caption{RD performance of a more complex ELIC model on KITTI. ``ELIC(PA)" represents a more complex version of the original ELIC model after expanding the channel number. The parameter size is essentially equivalent to that of ``ELIC-PC".}
  \label{fig:com_rd}
\end{figure}

To investigate the impact of parameter quantity on performance, we increase the parameters of the original image compression network. The augmentation involves simply expanding the channel number while keeping their parameters on par with the model using point cloud information. Following an identical training strategy, we train both models, and the rate-distortion performance results are depicted in Figure~\ref{fig:com_rd} and Table~\ref{tab:code}. ``ELIC(PA)" refers to the original ELIC model after parameter augmentation, with the increased parameters being equivalent to ELIC-PC. However, it does not utilize point cloud information for compression assistance. Taking ELIC as the anchor, ``ELIC(PA)" shows a performance improvement of 3.24\%, while ELIC-PC demonstrates a performance improvement of 6.05\%. This indicates that even with equal parameters and all additional parameters dedicated to image compression itself, utilizing point cloud information still yields better performance compared to not using point cloud information.

\begin{table}
\centering
  \caption{BD-Rate (\%) for PSNR (dB) of different models on KITTI. The anchor is ELIC\cite{he2022elic}. ``$\triangle$" represents the performance improvement compared to the original one. ``PC-Gain" represents the performance improvement when using point cloud information compared to using a tensor of all zeros. ``$\downarrow$" represents that the lower the metric, the better. }
  \label{tab:zeros}
  \begin{tabular}{lccc}
    \toprule
    Model&BD-Rate(\%) $\downarrow$&$\triangle \downarrow$ & PC-Gain $\downarrow$\\
    \midrule
    ELIC-PC & -6.05 & -6.05 & -3.72\\
    ELIC-PC(zeros) & -2.33 & -2.33 & -\\
    HYPER-PC & +28.31 & -15.67 & -8.12\\
    HYPER-PC(zeros) & +36.43 & -7.55 & -\\
    MLIC-PC & -4.31 & -2.80 & -2.35\\
    MLIC-PC(zeros) & -1.96 & -0.45 & -\\
  \bottomrule
\end{tabular}
\end{table}

\subsubsection{Impact of Model Structure}
To explore the impact of model structure on performance, we adopt a model structure that is identical to our proposed method, with the only difference being the use of an all-zero tensor in place of point cloud information as input to the model. ``zeros" represents the input of the point cloud branch is an all-zero tensor. As shown in Table~\ref{tab:zeros}, even with only a tensor of all zeros as input, our model already shows a certain degree of performance improvement compared to the original one. After using point cloud information, the performance is further enhanced, with the improvement on HYPER, ELIC and MLIC being 8.12\%, 3.72\% and 2.35\%, respectively. The above results indicate that the simpler the model, the greater the performance improvement after using point cloud information.

\begin{table}
\centering
  \caption{Parameter size and coding time. ``De-time" means decoding time. ``PC-time" means point cloud processing time.}
  \label{tab:freq}
  \begin{tabular}{lccc}
    \toprule
    Model&Params(M)&De-time(ms)&PC-time(ms)\\
    \midrule
    ELIC & 31.66 & 87.3 & -\\
    ELIC-PC & 87.39 & 86.2 & 6.4\\
    ELIC(PA) & 89.19 & 90.1 & - \\
    HYPER & 6.92 & 2.1 & -\\
    HYPER-PC & 28.20 & 3.1 & 5.1\\
    MLIC & 116.48 & 145.2 & -\\
    MLIC-PC & 174.06 & 147.6 & 4.8\\
  \bottomrule
\end{tabular}
\end{table}

\subsubsection{Parameter Size and Coding Time}
Since our method extracts dense features from sparse point cloud information, we add additional parameters to the original compression models. The increase in parameters mainly comes from two aspects: (a) the entire pipeline for extracting point cloud features and (b) the increased number of channels for a few convolutional layers to integrate point cloud information. The increased complexity is mainly in the point cloud branch and has minimal impact on the complexity of the compression model itself. As shown in Table~\ref{tab:freq}, although the model size has increased, the decoding time is roughly the same as the original model, and the time spent on the point cloud branch is minimal.

\section{Conclusion}
\label{sec:conclusion}

In this paper, we propose a method utilizing sparse point clouds to assist in image compression, which can be easily integrated into many existing learned frameworks to enhance compression performance in autonomous driving scenarios. We design PIP module to predict image information from point clouds, and the prediction results are input into MCM module to obtain multi-scale structural features. We further propose a general approach to integrate these features into compression networks. The experimental results demonstrate that our method not only improves performance but also better preserves structural information. While we achieve noticeable improvements across different frameworks, we believe there is still room for further enhancement. For instance, PIP and MCM may be able to be reused across different models at different bitrates, rather than retraining each time. We also expect to optimize the training strategy, deliberate the specific design of the modules, and simplify our model while maintaining performance through lightweight designs in our future work.

\ifCLASSOPTIONcaptionsoff
  \newpage
\fi

\bibliographystyle{ieeetr}
\bibliography{refs}
 
\begin{IEEEbiography}[{\includegraphics[width=1in,height=1.25in,clip,keepaspectratio]{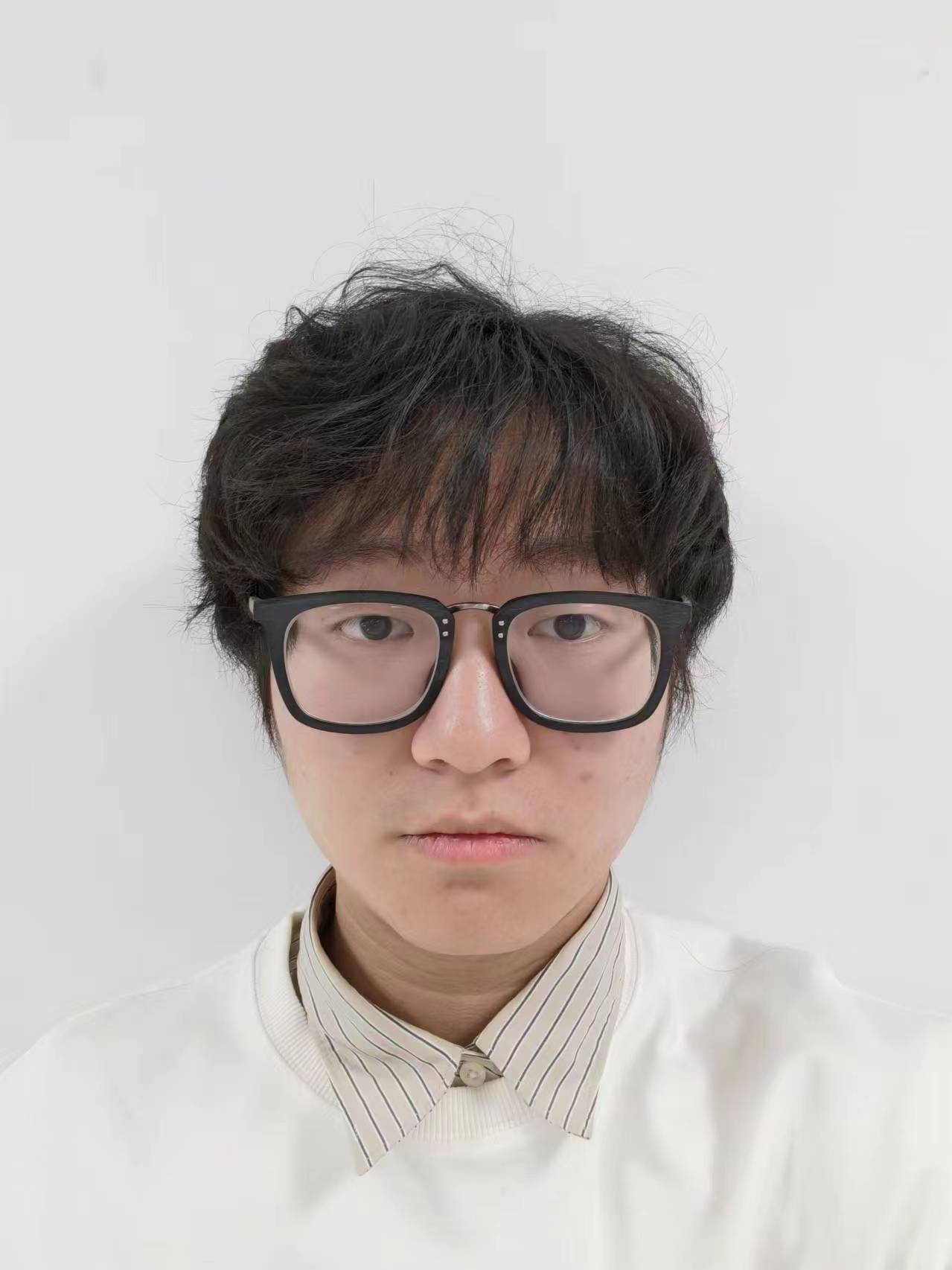}}] {Yiheng Jiang} received the B.S. degree in electronic information engineering from University of Science and Technology of China (USTC) in 2022.
He is currently pursuing the Ph.D. degree in the Department of Electronic Engineering and Information Science at USTC.
His research interests include point cloud/image/multi-modality coding and processing. 
\end{IEEEbiography}

\begin{IEEEbiography}[{\includegraphics[width=1in,height=1.25in,clip,keepaspectratio]{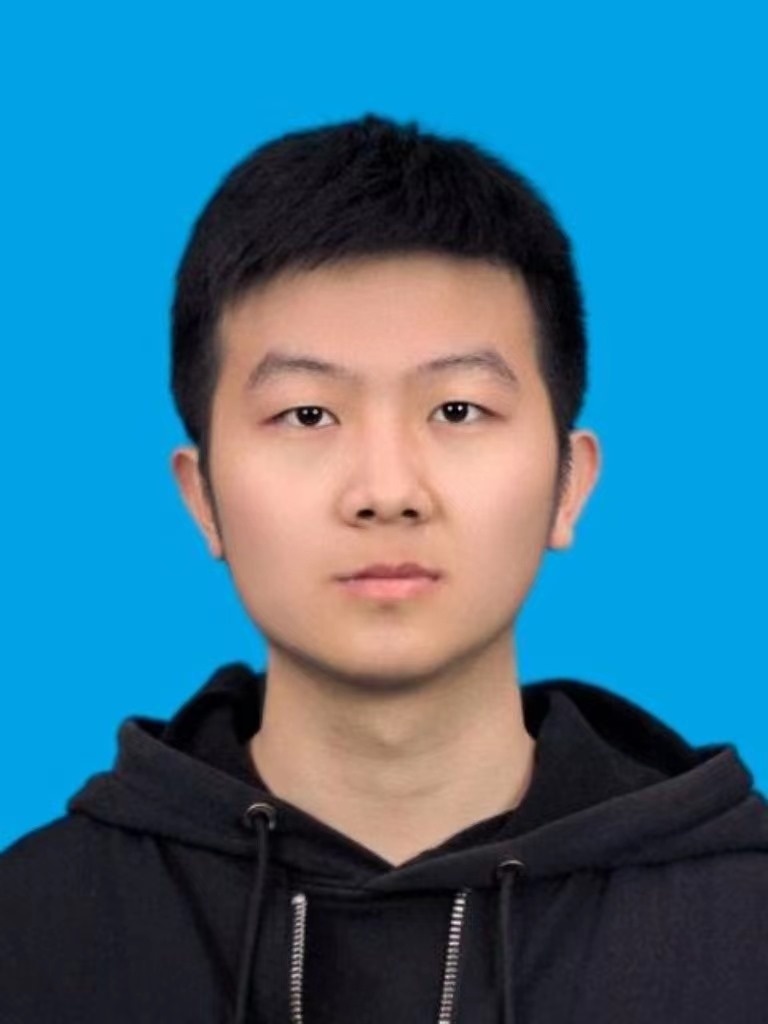}}]{Haotian Zhang}
received the B.S. degree in electronic information engineering from the University of Science and Technology of China (USTC), Hefei, China, in 2021. He is currently pursuing the Ph.D. degree in the Department of Electronic Engineering and Information Science at USTC. His research interests include image/video coding, machine learning, and computer vision.
\end{IEEEbiography}

\begin{IEEEbiography}[{\includegraphics[width=1in,height=1.25in,clip,keepaspectratio]{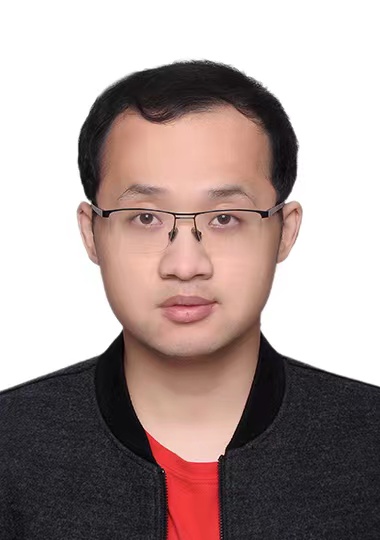}}] {Li Li} (M'17) received the B.S. and Ph.D. degrees in electronic engineering from University of Science and Technology of China (USTC), Hefei, Anhui, China, in 2011 and 2016, respectively.
He was a visiting assistant professor in University of Missouri-Kansas City from 2016 to 2020.
He joined the department of electronic engineering and information science of USTC as a research fellow in 2020 and became a professor in 2022.

His research interests include image/video/point cloud coding and processing.
He has authored or co-authored more than 80 papers in international journals and conferences. 
He has more than 20 granted patents. 
He has several technique proposals adopted by standardization groups.
He received the Multimedia Rising Star 2023.
He received the Best 10\% Paper Award at the 2016 IEEE Visual Communications and Image Processing (VCIP) and the 2019 IEEE International Conference on Image Processing (ICIP).
He serves as an associate editor for \textsc{IEEE Transactions on Circuits and Systems for Video Technology} from 2024 to 2025. 
\end{IEEEbiography}

\begin{IEEEbiography}[{\includegraphics[width=1in,height=1.25in,clip,keepaspectratio]{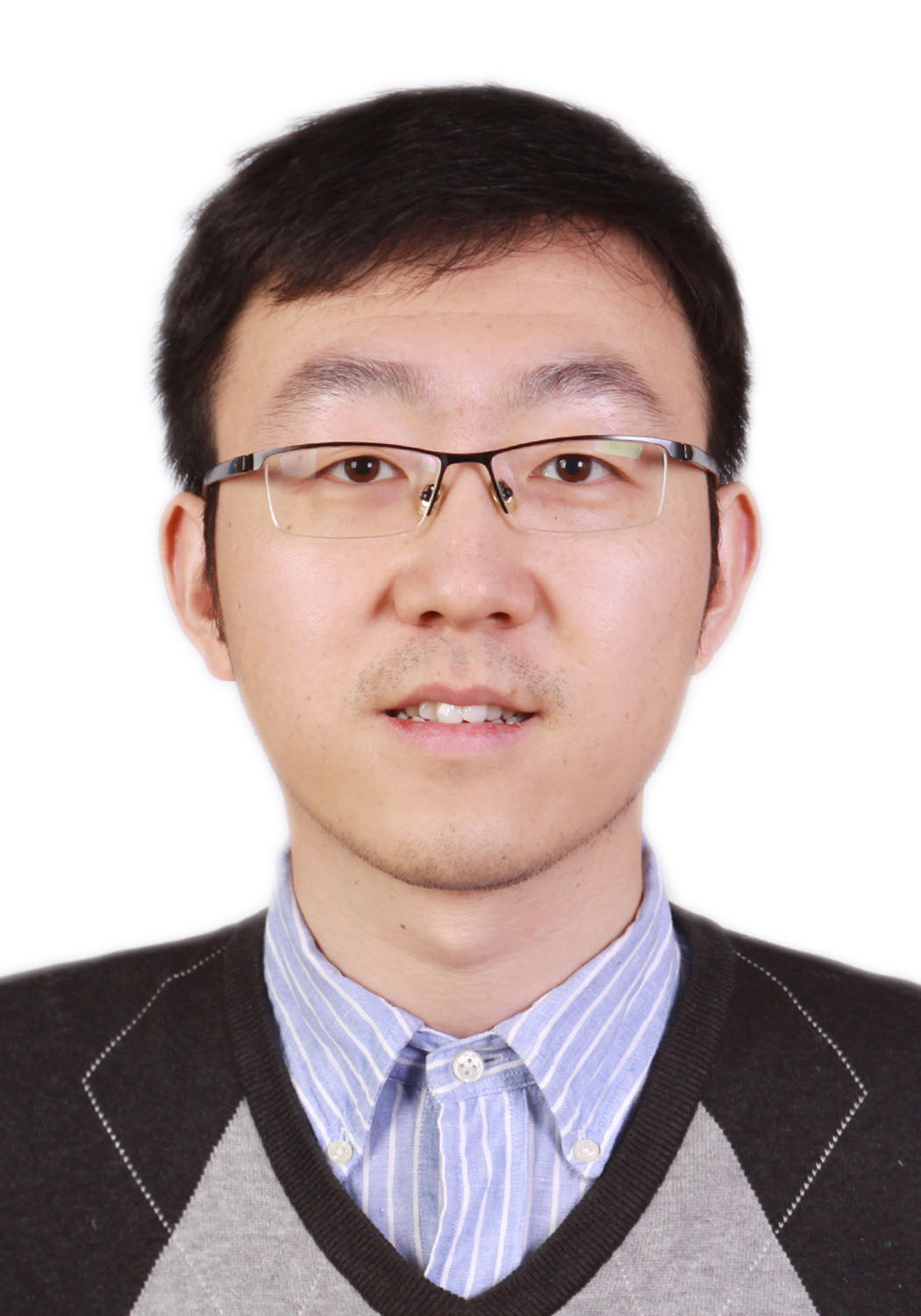}}]{Dong Liu}
(M'13--SM'19) received the B.S. and Ph.D. degrees in electrical engineering from the University of Science and Technology of China (USTC), Hefei, China, in 2004 and 2009, respectively. He was a Member of Research Staff with Nokia Research Center, Beijing, China, from 2009 to 2012. He joined USTC as a faculty member in 2012 and became a Professor in 2020.

His research interests include image and video processing, coding, analysis, and data mining.
He has authored or co-authored more than 200 papers in international journals and conferences. He has more than 30 granted patents. He has several technique proposals adopted by standardization groups.
He received the 2009 \textsc{IEEE Transactions on Circuits and Systems for Video Technology} Best Paper Award, VCIP 2016 Best 10\% Paper Award, and ISCAS 2022 Grand Challenge Top Creativity Paper Award. He and his students were winners of several technical challenges held in ISCAS 2023, ICCV 2019, ACM MM 2019, ACM MM 2018, ECCV 2018, CVPR 2018, and ICME 2016. He is a Senior Member of CCF and CSIG, and an elected member of MSA-TC of IEEE CAS Society. He serves or had served as the Chair of IEEE 1857.11 Standard Working Subgroup (also known as Future Video Coding Study Group), an Associate Editor for \textsc{IEEE Transactions on Image Processing}, a Guest Editor for \textsc{IEEE Transactions on Circuits and Systems for Video Technology}, an Organizing Committee member for VCIP 2022, ChinaMM 2022, ICME 2021, etc.
\end{IEEEbiography}

\begin{IEEEbiography}[{\includegraphics[width=1in,height=1.25in,clip,keepaspectratio]{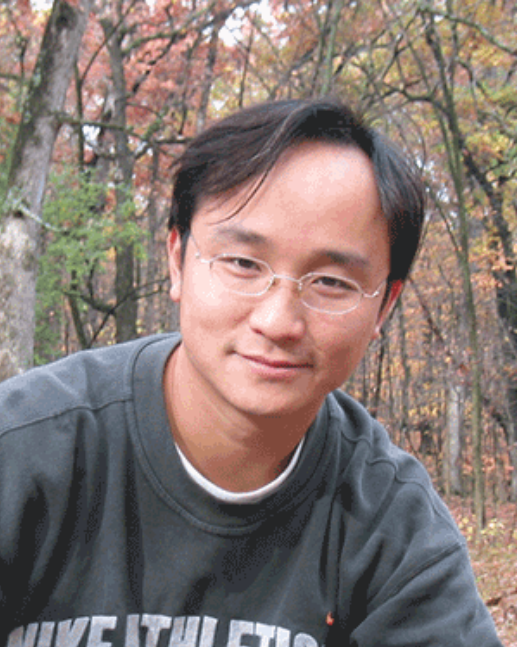}}]{Zhu Li}
Zhu Li is a full professor with the Dept of Computer Science \& Electrical Engineering, University of Missouri, Kansas City(UMKC), and the director of NSF I/UCRC Center for Big Learning (CBL) at UMKC. He received his PhD in Electrical \& Computer Engineering from Northwestern University in 2004. He was AFRL summer faculty at the UAV Research Center, US Air Force Academy (USAFA), 2016-18, 2020-23. He was Senior Staff Researcher with the Samsung Research America's Multimedia Standards Research Lab in Richardson, TX, 2012-2015, Senior Staff Researcher with FutureWei Technology's Media Lab in Bridgewater, NJ, 2010~2012, Assistant Professor with the Dept of Computing, the Hong Kong Polytechnic University from 2008 to 2010, and a Principal Staff Research Engineer with the Multimedia Research Lab (MRL), Motorola Labs, from 2000 to 2008. His research interests include point cloud and light field compression, graph signal processing and deep learning in the next gen visual compression, remote sensing, image processing and understanding. He has 50+ issued or pending patents, 200+ publications in book chapters, journals, and conferences in these areas. He is an IEEE senior member, Associate Editor-in-Chief for IEEE Trans on Circuits \& System for Video Tech, 2020~, Associate Editor for IEEE Trans on Image Processing(2020~), IEEE Trans.on Multimedia (2015-18), IEEE Trans on Circuits \& System for Video Technology(2016-19). He received the Best Paper Runner-up Award at the Perception Beyond Visual Spectrum (PBVS) grand challenge at CVPR 2023, Best Paper Award at IEEE Int'l Conf on Multimedia \& Expo (ICME), Toronto, 2006, and IEEE Int'l Conf on Image Processing (ICIP), San Antonio, 2007.
\end{IEEEbiography}

\end{document}